%% file: main.tex
\documentclass{article}
\usepackage{amssymb}
\usepackage{booktabs}
\usepackage{siunitx}
\usepackage{amsmath} 
\usepackage{bm}
\usepackage{authblk} 
\usepackage{graphicx}
\usepackage{xcolor}
\usepackage{graphicx}
\usepackage{wrapfig}
\usepackage{hyperref}
\usepackage[space, compress, sort]{cite}

\usepackage[preprint]{corl_2025} 

\title{LLM-Land: Large Language Models for Context-Aware Drone Landing}

\author{
  Siwei Cai\textsuperscript{1},
  Yuwei Wu\textsuperscript{2},
  Lifeng Zhou\textsuperscript{1}\thanks{Corresponding author}\\
   \textsuperscript{1}Department of Electrical and Computer Engineering \\
  Drexel University,
  United States\\
   \textsuperscript{2}Department of Electrical and Systems Engineering \\
  University of Pennsylvania,
  United States\\
  \texttt{\{sc3568, lz457\}@drexel.edu}, \quad \texttt{yuweiwu@seas.upenn.edu}
}

\begin{document}
\maketitle

\begin{abstract} 
\input{sections/0_abstract}
\end{abstract}



\section{Introduction}
\label{sec:introduction}
\input{sections/1_introduction}


\section{Related Works}
\label{sec:related_works}
\input{sections/2_related_works}
	

\section{Methods}
\label{sec:methods}
\input{sections/3_methods}


\section{Results}
\label{sec:results}
\input{sections/4_results}


\section{Conclusion and Discussion}
\label{sec:conclusion}
\input{sections/5_conclusion_and_discussion}

\clearpage
\section{Limitations}
Despite these promising results, our system exhibits several limitations. First, the perception pipeline relies solely on image captions to supply semantic context, but does not embed precise obstacle geometry or spatial relationships within the LLM’s reasoning process—a shortcoming that could be addressed by adopting a more powerful vision encoder such as DAM-3B \cite{lian2025describe}. Second, we face limitations stemming from the intrinsic capabilities of the lightweight LLMs suitable for onboard deployment. These models, necessarily smaller due to resource constraints, lack the parameter scale and architectural complexity of leading large-scale models, resulting in shallower logical reasoning and a reduced capacity for complex inference. This gap limits our ability to tackle tasks requiring deep semantic understanding or intricate planning, highlighting the need for more capable model architectures feasible for edge deployment. 
Third, beyond the model's inherent capacity, the operational speed on current embedded hardware presents a significant bottleneck. Real-time autonomy demands low-latency decision-making, but the computational constraints of typical onboard processors lead to slow LLM inference. This high latency restricts the practical use of complex reasoning, as generating timely responses for dynamic situations like landing becomes infeasible. Thus, improvements in hardware processing power and efficiency are crucial to enable the execution of more sophisticated reasoning tasks within critical time constraints.

\acknowledgments{If a paper is accepted, the final camera-ready version will (and probably should) include acknowledgments. All acknowledgments go at the end of the paper, including thanks to reviewers who gave useful comments, to colleagues who contributed to the ideas, and to funding agencies and corporate sponsors that provided financial support.}


\bibliography{reference}  


\newpage
\appendix
\input{sections/6_appendix}


\end{document}

%% file: sections/0_abstract.tex
Autonomous landing is essential for drones deployed in emergency deliveries, post-disaster response, and other large-scale missions. By enabling self-docking on charging platforms, it facilitates continuous operation and significantly extends mission endurance. However, traditional approaches often fall short in dynamic, unstructured environments due to limited semantic awareness and reliance on fixed, context-insensitive safety margins. To address these limitations, we propose a hybrid framework that integrates large language model (LLMs) with model predictive control (MPC). Our approach begins with a vision–language encoder (VLE) (e.g., BLIP), which transforms real-time images into concise textual scene descriptions. These descriptions are processed by a lightweight LLM (e.g., Qwen 2.5 1.5B  or LLaMA 3.2 1B) equipped with retrieval-augmented generation (RAG) to classify scene elements and infer context-aware safety buffers, such as 3 meters for pedestrians and 5 meters for vehicles. The resulting semantic flags and unsafe regions are then fed into an MPC module, enabling real-time trajectory replanning that avoids collisions while maintaining high landing precision. We validate our framework in the ROS-Gazebo simulator, where it consistently outperforms conventional vision-based MPC baselines. Our results show a significant reduction in near-miss incidents with dynamic obstacles, while preserving accurate landings in cluttered environments.

\vspace{0.5em}
\noindent\textbf{Supplementary video:} \href{https://youtu.be/9yGEpqmCtdA}{https://youtu.be/9yGEpqmCtdA}

%% file: sections/1_introduction.tex


Unmanned Aerial Vehicles (UAVs) have become indispensable across a wide range of applications, including surveillance \cite{kingston2008decentralized}, package delivery \cite{saunders2024autonomous}, infrastructure inspection \cite{aela2024uav}, environmental monitoring \cite{fang2024strategies}, and precision agriculture \cite{toscano2024unmanned}. Landing represents one of the most critical and technically demanding phases of UAV operation in these applications
\cite{6842377, Alam2021ASOA,Xin2022VisionBasedALA}. A safe, autonomous landing capability not only guarantees mission completion but also unlocks advanced workflows such as recharging or payload handoff, which significantly improve UAV endurance and flexibility \cite{cai2023energy,bin2017autonomous}. 
Delivering this capability within the strict limits of UAV platforms, which are characterized by limited battery capacity and restricted onboard computational resources, requires autonomous landing systems that are both highly efficient and reliably robust~\cite{Alsamhi2022ComputingITA,Yu2021ECSAGINsESA}.

The primary challenge in autonomous landing arises in dynamic, uncertain settings where conventional navigation strategies often prove inadequate. 
While such methods are effective at optimizing trajectories within accurately modeled systems \cite{Hanover2021PerformancePAA, Ba2019AutonomousLOA, liu2017ral}, they rely on predefined environmental models and typically lack the adaptability required to handle unforeseen variations or rapidly changing conditions in the landing scene. 
However, real-world landing settings are highly variable, ranging from unstructured natural terrains such as forests and waterfronts to dense urban landscapes populated with static obstacles like buildings and vehicles, as well as dynamic entities such as pedestrians or animals \cite{Mittal2018VisionbasedALA}. 
Successfully managing this diversity requires approaches that are not only robust to environmental variability but also capable of context-aware reasoning in real time.
A further limitation of conventional landing pipelines lies in their reliance on raw sensor data, which typically lacks semantic understanding~\cite{10802806}.
Standard perception stacks can detect an object's presence and location, but often cannot ascertain its identity, such as distinguishing an inert rock from a living being. 
This semantic gap is crucial: a rock or other inanimate object might only need a small detour, but a person or any vulnerable entity calls for an immediate, more dramatic avoidance maneuver. 
Closing this gap requires more than refining detection and planning modules to fit specific scenarios, it requires a system capable of recognizing the semantic nature of obstacles and reasoning about their potential behavior to support safe and context-aware landing decisions.

\begin{figure}[!t]
    \centering
    \includegraphics[width=\linewidth]{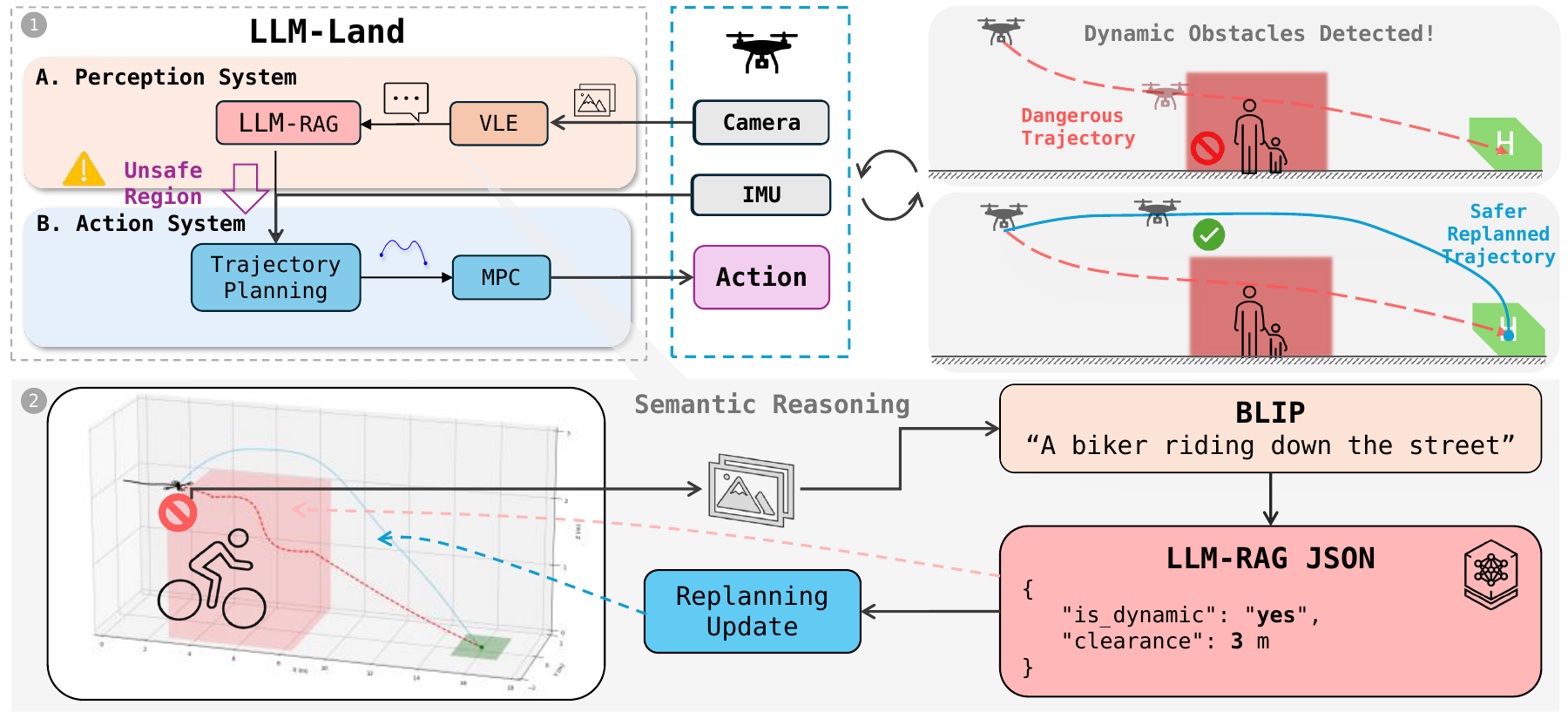}
    \caption{Overview of the LLM-Land framework for semantics-driven safe landing.
    (A) The Perception System receives raw images from the onboard camera and converts them into scene captions using a vision–language pretrained encoder model (e.g., BLIP). An LLM augmented with Retrieval-Augmented Generation (RAG) analyzes each caption to detect dynamic obstacles and output the required unsafe region, emitting a structured JSON specification.
    (B) The Action System ingests the JSON input and collaborates with a local MPC-based planner to generate actions for a safe landing. Top-right illustration shows the landing action sequence. Bottom panel detailed semantic reasoning process.}
    \label{fig: TopOverview}
    \vspace{-0.5cm}
\end{figure}

To address this challenge, we propose leveraging the abstraction and reasoning capabilities inherent in Large Language Models (LLMs) \cite{li2025large, Yue2025ASOA}. Specifically, we employ an LLM as a high-level reasoning engine capable of interpreting complex environmental contexts, characterizing diverse obstacle types, and making informed decisions to guide the landing process with greater intelligence and safety.

However, integrating LLMs into the real-time, safety-critical control loop of a UAV presents several specific challenges. Firstly, LLMs operate on processed information, typically text. To enable the LLM to ``perceive" the environment, visual data captured by the UAV's onboard cameras must be translated into a format the LLM can understand. We achieve this by incorporating a vision-language encoder (VLE) such as CLIP~\cite{radford2021learning}, BLIP~\cite{li2022blip}, or BLIP-2~\cite{li2023blip}, which processes sensor data (primarily images) and extracts relevant features or descriptions to be fed into the LLM.

Second, in safety-critical applications like drone landing, ensuring the reliability and factual accuracy of the LLM's output is crucial. Lightweight LLMs are better suited for deployment on resource-constrained UAV hardware. However, these models can sometimes be unstable as they may produce ``hallucinations" which are responses that sound plausible but are factually wrong or nonsensical in context \cite{Sun2025OnionEvalAUA}. To mitigate this risk and improve the dependability of the LLM's reasoning, we employ a Retrieval-Augmented Generation (RAG) mechanism \cite{Lewis2020RetrievalAugmentedGFA, Gupta2024ACSA}. The RAG framework guides the LLM's output generation process by dynamically retrieving relevant information from a curated knowledge base (containing safe operating procedures, obstacle characteristics, etc.), ensuring that the LLM's directives are contextually appropriate and grounded in verified data \cite{Ram2023InContextRLA}. 

Thirdly, translating the LLM’s high-level reasoning into concrete UAV commands requires a hybrid approach. Rather than depending solely on LLM to generate the required actions, we embed the LLM’s guidance into a conventional planning‐and‐control pipeline that integrates dynamically-feasible trajectory search with an MPC-based trajectory tracker for robust performance over extended missions with formal guarantees~\cite{BANGURA201411773}. In this hybrid scheme, the LLM acts as a supervisory layer: it refines safety constraints and strategic objectives (for example, by dynamically adjusting MPC parameters or defining adaptive unsafe regions around sensitive areas), while the underlying planning framework handles robust, real-time trajectory optimization and execution.

Finally, the entire system must be computationally efficient enough to operate in real-time on the UAV's limited hardware, enabling interventions at a sufficiently high frequency to react effectively to dynamic events during landing. This necessitates the use of lightweight and optimized models for both the VLE and LLM components, along with streamlined communication pathways between the perception (VLE), reasoning (LLM+RAG), and planning (MPC) modules.

Specifically, this paper addresses the challenging scenario where a UAV is required to land under diverse and dynamic environmental conditions. Our proposed framework (Fig.~\ref{fig: TopOverview}) combines a VLE, a lightweight LLM stabilized by RAG, and a local planner with MPC modulated by the LLM's reasoning to provide robust, adaptive, and safe autonomous landing capabilities.

Our main contributions are threefold:
\begin{itemize}
\item We introduce a novel approach that leverages the VLE's perception and the LLM's reasoning to endow a UAV with an understanding of various environmental obstacles, assisting the UAV in making smarter, safer landing choices based on the situation.
\item We propose and demonstrate a method for stabilizing lightweight LLMs for safety-critical UAV landing using a RAG mechanism, achieving reliable reasoning performance on resource-constrained platforms. 
\item We present an integrated LLM-MPC control framework capable of high-frequency, real-time operation, effectively bridging high-level semantic reasoning with low-level motion control for autonomous UAV landing. 
\end{itemize}

%% file: sections/2_related_works.tex
UAV autonomous landing has been extensively investigated. Niu et al. \cite{niu2021vision} propose a vision-guided approach for cooperative UAV–UGV landing by placing multiple QR codes on the UGV platform. The relative pose, speed, and heading are obtained from QR-code–based vision combined with either VIO or GPS. These measurements are then fed into a velocity controller that integrates control-barrier and control-Lyapunov functions to ensure a safe descent onto a moving target. Zhang et al.~\cite{zhang2023coni} present a concise target-frame MPC that relies only on relative pose, velocity, and IMU data (no global state estimation or motion models) to achieve centimeter-level landings on dynamic bases, validated both indoors (motion capture) and outdoors (onboard sensing). While these methods attain impressive precision in structured, obstacle-free settings, they generally assume an empty testbed (no static obstacles, let alone dynamical hazards) and thus overlook the complexity of real‐world landing zones populated by people, vehicles, or debris.

Beyond the UAV-landing literature, the broader robotics community has developed rich environment‐understanding pipelines and begun to explore LLM‐based control. Classical perception methods such as semantic segmentation, occupancy grids or SLAM provide detailed maps but rarely include explicit obstacle classification, so they cannot reliably distinguish critical hazards (e.g. humans) from benign terrain. Wang et al.~\cite{wang2025contact} introduce CAMP, a contact-aware motion planner that embeds complementarity constraints into an augmented Lagrangian methods optimization to handle intentional object interactions, but it does not continuously fuse live visual observations for onboard reactivity. R. Sinha et al.~\cite{sinha2024real} propose a two-stage LLM-powered runtime monitor: a fast binary classifier in embedding space flags out-of-distribution anomalies, then a generative LLM fallback verifies and guides an MPC‐based safe fallback trajectory. Although their hybrid scheme boosts safety under anomalous conditions, the UAV must hover while waiting for the slower LLM response, degrading operational efficiency.

In contrast, our approach tightly couples high-frequency, vision-based obstacle classification with triggered trajectory replanning. By directly observing the environment at control rates, we integrate perception and action to address both static and dynamic obstacles during descent. This allows real-time updates on safety buffers and flight paths without incurring the latency of repeated large-model inference. This unified pipeline brings landing research closer to the demands of cluttered, unpredictable real‐world scenarios.




%% file: sections/3_methods.tex


\paragraph{Overall System Architecture.} We tackle the problem of context-aware autonomous landing of a UAV in dynamic, unstructured environments by decomposing our solution into two tightly coupled subsystems: \emph{Perception} and \emph{Action} (Fig.~\ref{fig: TopOverview}). The Perception subsystem converts raw images into structured safety parameters via a VLE and a lightweight LLM with RAG. The Action subsystem then consumes these parameters in a 3D MPC-based planning framework to generate and execute collision-free landing trajectories in real time.

\subsection{Perception System}
The perception subsystem translates pixel streams into JSON–formatted safety constraints. It comprises two sequential stages: 
1) \emph{Scene Understanding}, which generates free-form captions of the environment via a VLE, and 
2) \emph{Semantic Reasoning}, which converts those captions into precise obstacle classifications and unsafe regions using an LLM augmented with RAG.

\textbf{Scene Understanding via VLE.}\label{sec:scene-understanding}
A node subscribes to the forward‐facing camera, always retaining the latest frame. Each frame is resized to BLIP’s required input resolution (384×384 px) and passed to the BLIP (V1) model~\cite{li2022blip}, which uses a Vision Transformer encoder plus an auto-regressive decoder to produce a concise, zero-shot caption.
BLIP is chosen over contrastive models such as CLIP \cite{radford2021learning} because its generative decoder requires no predefined prompt set. We use the original BLIP  rather than BLIP-2 \cite{li2023blip} to minimize on-board memory and computation overhead, which is a crucial consideration for resource-limited UAVs.

\textbf{Semantic Reasoning via LLM–RAG.}\label{sec:semantic-reasoning}  
Captions from BLIP enter a four-step pipeline that outputs exactly two fields in JSON. 
\begin{itemize}
   \item  \emph{Knowledge Base \& Indexing.}  
    We maintain a JSON-structured onboard knowledge base (KB) encoding: obstacle classes (static vs dynamic, human vs inanimate), minimum safe altitudes, and nominal safety buffers. Documents are chunked into 512-token segments and indexed using LlamaIndex’s VectorStoreIndex~\cite{Liu_LlamaIndex_2022} with BAAI/bge-small-en-v1.5 embeddings~\cite{bge_embedding}. We omit re-ranking for lower retrieval latency, but the architecture allows a later plug-in of a learned re-ranker if KB grows.
    \item \emph{Retrieval.}  
    Each BLIP caption is embedded with the same BAAI/bge-small-en-v1.5 embeddings and queried in hybrid mode (semantic + keyword). Hybrid ranking merges dense-vector similarity with inverted-index term matches to balance conceptual relevance and exact keyword hits. We retrieve the top three segments to supply the LLM with enough context without exceeding our real-time deadline.
   \item  \emph{LLM Inference \& Optimization.}  
    We run a 4-bit GPTQ-quantized, instruction-tuned LLM (e.g.\ \texttt{Llama3.2-1B-Instruct} or \texttt{Qwen2-1.5B-Instruct}) via Unsloth~\cite{han2023unsloth}. Static KV caching plus quantization reduces VRAM and doubles throughput versus a vanilla Hugging Face pipeline. We enforce determinism (temperature \(\tau\leq0.5\)), cap outputs at 20 tokens, and enable early stopping on end-of-sequence to guarantee bounded latency.
    \item \emph{Prompting \& Publishing.}  A minimal system prompt forces the LLM to emit only the two JSON fields shown in part 2 of Fig~\ref{fig: TopOverview}. After generation, we parse \texttt{is\_dynamic} and \texttt{z\_min} and publish them on a dedicated ROS topic. The downstream MPC planner then uses these values to adjust the landing trajectory and enforce the correct safety buffer and altitude floor.
\end{itemize}


\subsection{Action System}

When feedback is received from the perception system, a motion planning and trajectory tracking pipeline with MPC is deployed to enable safe and feasible control actions. 
We follow the framework proposed in~\cite{9531427}, which combines a trajectory search strategy with a nonlinear MPC to track a dynamically feasible reference trajectory.

\textbf{3-D Nonlinear MPC.}
The nonlinear system state is defined as \( \bm{x} = [\bm{p}, \bm{v}, \phi, \theta, \psi]^{\top} \in \mathcal{X} \), where \( \bm{p}, \bm{v} \in \mathbb{R}^{3} \) denote the position and velocity, and \( \phi, \theta, \psi \in \mathbb{R} \) represent roll, pitch, and yaw angles. 
The control input is \( \bm{u} = [\dot{\phi}_{c}, \dot{\theta}_{c}, \dot{\psi}_{c}, T_{c}]^{\top} \in \mathcal{U} \), as commanded angular rates and thrust.
With a reference trajectory \( \{ \bm{x}_{\text{ref}}^{k} \}_{k=0}^{N} \), the MPC problem over a finite horizon \(N\) that satisfies the system dynamics and collision-free constraints can be formulated by
\begin{subequations}
\begin{align}
    &\min_{\bm{x}, \bm{u}} J^{N}(\bm{x}^{N}, \bm{x}_{\text{ref}}^{N}) + \sum_{k=0}^{N-1} \left( J_{\bm{x}}^{k}(\bm{x}^{k}, \bm{x}_{\text{ref}}^{k}) + J_{\bm{u}}^{k}(\bm{u}^{k}) + J_{\Delta \bm{u}}^{k}(\bm{u}^{k+1}, \bm{u}^{k}) \right), \\
    \text{s.t.} \quad & \bm{x}^{k+1} = f(\bm{x}^{k}, \bm{u}^{k}),  \ \bm{x}^{0} = \bm{x}_0, \\
    & \bm{u}^{k} \in \mathcal{U}, \quad \bm{x}^{k} \in \mathcal{X}, \quad \bm{p}^{k} \in \mathcal{P}_c ,
\end{align}
\end{subequations}
where \( \bm{x}^{k} \) and \( \bm{u}^{k} \) denote the system state and control input at time step \(k\), and $f$ represents the nonlinear system dynamics. The objective penalizes state tracking error, control effort, and input variation, respectively.
Each reference state \( \bm{x}_{\text{ref}}^{k} \) includes only the desired position and orientation components.  
Velocity and higher-order derivatives are not explicitly provided in the reference and are instead regulated by the MPC optimization to ensure dynamic feasibility.
The collision-free constraints are represented using convex polytopes $\mathcal{P}_c = \{ x \in \mathbb{R}^3 \mid \bm{A}_c x \leq \bm{b}_c \}$, and integrated as linear inequalities and dynamically updated during planning.
The resulting trajectories are fully executable by the low-level controller, such as an \(SE(3)\) geometric controller~\cite{5717652}.

\textbf{Context-Aware Real-Time Trajectory Update.} When a potential dynamic obstacle enters the scene, the system initiates a real-time replanning process by updating the front-end trajectory search and corridor constraints. 
The LLM-RAG module identifies unsafe regions by reasoning about the dynamic obstacle’s intent and outputs these regions to the action system. 
The unsafe region associated with the dynamic obstacle is modeled as another convex set $\mathcal{B}$.
To ensure obstacle avoidance, the resulting feasible space must simultaneously satisfy the original corridor constraint and exclude the interior of $\mathcal{B}$. 
During the trajectory search phase, explicit collision checking is applied to candidate paths against the identified unsafe regions, allowing the system to discard trajectories that intersect with unsafe regions. In the MPC module, we refine the convex polytopes used in the constraint formulation to explicitly exclude these unsafe regions. Specifically, we construct a modified feasible set \(\tilde{\mathcal{P}}_c\) such that \(\tilde{\mathcal{P}}_c \subseteq \mathcal{P}_c\) and \(\tilde{\mathcal{P}}_c \cap \mathcal{B} = \emptyset\), ensuring that the final trajectories are confined to a safe subset of the original corridor.

%% file: sections/4_results.tex


To validate the effectiveness of our proposed LLM-Land framework for autonomous UAV landing, we conducted extensive simulations to compare its performance against baseline methods quantitatively. Additionally, we carried out preliminary real-world tests to assess its feasibility. Additional experimental results are provided in the Appendix, and the accompanying video is available as supplementary material.


\subsection{Experimental Setup}\label{subsec:experimental_setup} 

\noindent \textbf{Simulation Environment.}  
All experiments were implemented in ROS Noetic, integrated with the Gazebo simulator via the \texttt{rotors\_sim} package~\cite{Furrer2016}. This setup provides realistic physics and sensor modeling, including RGB-D camera data. 

\noindent \textbf{Hardware Configuration.}  
Simulations ran on a laptop equipped with an Intel Core i7-12800H CPU, 16 GB of system RAM, and an NVIDIA GeForce RTX 3070 Ti Mobile GPU (8 GB VRAM). This configuration reflects realistic onboard constraints, particularly the GPU VRAM limit, which is critical for evaluating model efficiency.

\begin{figure}[!t]
    \centering
    \includegraphics[width=\linewidth]{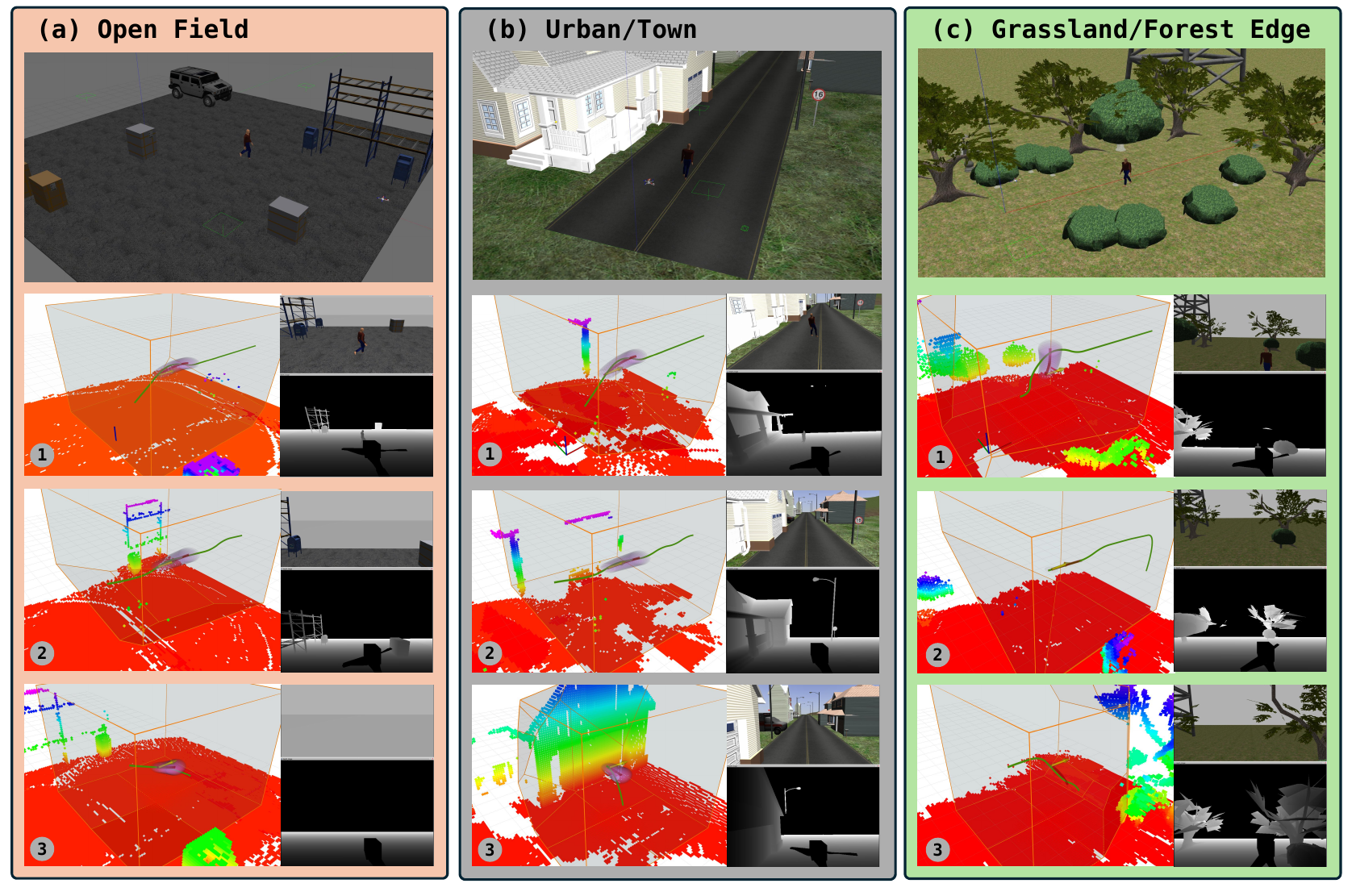}
    \caption{Experiments were conducted using the Gazebo simulator, which provides realistic physics and sensor modeling, including support for both depth and RGB-D cameras. We evaluated LLM-Land across three representative scenarios:
(a) Open Field: a large, flat area with minimal static obstacles;
(b) Urban / Town: a built-up environment featuring buildings, streets, and parked vehicles;
(c) Grassland / Forest Edge: a semi-natural setting with outdoor structures, trees, and underbrush.}
    \label{fig: testenv}
    \vspace{-0.5cm}
\end{figure}

We evaluated performance across three distinct scenarios as shown in Fig.~\ref {fig: testenv}: (a) \textbf{Open Field:} A large, flat area with minimal static obstacles; (b) \textbf{Urban / Town:} A cityscape featuring buildings, streets, static vehicles; (c) \textbf{Grassland / Forest Edge:} A semi-natural environment with terrain, trees, and bushes.

\subsection{Compared Methods}
\label{subsec:compared_methods}

We benchmarked three distinct landing methods in our experiments: (a) \textbf{Baseline MPC}, a standard model predictive controller designed for UAV landing that relies solely on geometric obstacle information from a depth sensor, without any semantic reasoning; (b) \textbf{LLM–MPC}, which augments the baseline MPC with a lightweight LLM that processes scene captions generated by BLIP to provide high-level guidance—without RAG; and (c) \textbf{LLM-Land}, our proposed framework that incorporates RAG into the LLM pipeline, enabling stabilized, context-aware reasoning to inform and guide the MPC.

\subsection{Experiment 1: Dynamic Obstacle Avoidance During Landing}
\label{subsec:exp1_dynamic_obstacle}

As shown in Fig.~\ref {fig: testenv} (a), we tested the UAV on a large, flat area with minimal static obstacles. The UAV begins 2 meters above and laterally offsets from the target landing zone. During the landing process, a simulated humanoid figure moves unpredictably near the path to the landing area.

\textbf{Metric.}  
We classify a trial as successful when the UAV touches down within a 0.5 m radius of the target without colliding with the dynamic obstacle or entering its 1 m safety buffer. Approaches that breach this 1 m safety buffer are classified as close calls, which, in a human context, would be dangerously close to a person’s head or other body parts. Each method was evaluated over 50 independent trials.

\sisetup{
  detect-weight   = true,      
  table-format    = 3.0,       
  table-space-text-post = {\%} 
}
\begin{table}[h!]
  \centering
  \caption{Landing Success Rate with Dynamic Obstacle Avoidance (50 Trials)}
  \label{tab:exp1_dynamic_obstacle}
  \begin{tabular}{@{}l S[table-format=3.0]@{}}
    \toprule
    \textbf{Method} &
    \multicolumn{1}{c}{\textbf{Success Rate (\%)}} \\
    \midrule
    Baseline MPC    &  34.0   \\
    LLM–MPC         &   6.0  \\
    LLM-Land     &  \bfseries 96.0 \\
    \bottomrule
  \end{tabular}
\end{table}

\textbf{Results.}  
Tab.~\ref{tab:exp1_dynamic_obstacle} reports the landing success rates. 
As observed, the baseline MPC achieves a mere 34\% success rate in the presence of dynamic obstacles. When augmented with an LLM alone, performance further degrades to just 6\%. We attribute this decline to the LLM’s inconsistent and semantically incoherent outputs, which disrupt the MPC’s control logic rather than enhancing it.
Our LLM-Land framework attains the highest success rate ({\bfseries 96}\%), demonstrating the value of grounded, context-aware reasoning for reliable obstacle avoidance in the landing task.

\subsection{Experiment 2: Performance Across Diverse Environments}
\label{subsec:exp2_environments}
 
In Fig.~\ref {fig: testenv}, the drone with each landing method must autonomously land at a designated target in three environments: (a) Open Field: as  in experiment ~\ref{tab:exp1_dynamic_obstacle}; (b) Urban / Town: a built-up environment featuring buildings, streets, and parked vehicles;
(c) Grassland / Forest Edge: a semi-natural setting with outdoor structures, trees, and underbrush. 

\textbf{Metric.} We use the same metric for success rate as in Experiment~\ref{tab:exp1_dynamic_obstacle}.

\begin{table}[h!]
  \centering
  \caption{Landing Success Rate (\%) Across Diverse Simulation Environments}
  \label{tab:exp2_environments}
  \begin{tabular}{@{}l *{4}{S[table-format=2.1]}@{}}
    \toprule
    \textbf{Method}                   & \textbf{Open Field} & \textbf{Urban / Town}  & \textbf{Grassland / Forest} \\
    \midrule
    Baseline MPC                      & 34.0             & 28.0      & 32.0                \\
    LLM–MPC                           & 8.0            & 12.0      & 6.0                  \\
    LLM-Land              & \bfseries 94.0   & \bfseries 74.0      & \bfseries 80.0        \\
    \bottomrule
  \end{tabular}
\end{table}

\textbf{Results.}  
Tab.~\ref{tab:exp2_environments} summarizes the success rates. The Baseline MPC controller achieves only moderate success in the simplest environment (34\% in the Open Field) and degrades further in more cluttered settings (28\% in Urban, 32\% in Grassland/Forest). Augmenting MPC with just a lightweight LLM (LLM–MPC) did not improve its success rate: pervasive semantic “hallucinations” lead to uniformly poor performance (8\% –12\%), regardless of scenario. In contrast, our LLM-Land framework maintains consistently high landing success—94\% in the Open Field, 74\% in Urban, and 80\% in Grassland/Forest—demonstrating both robust semantic grounding and adaptability to diverse environmental complexities.


\subsection{Experiment 3: LLM Performance Benchmark on Constrained Hardware}
This experiment focused on the performance trade-offs of different lightweight LLMs integrated into our LLM+RAG-MPC framework, specifically considering the 8GB VRAM constraint of our simulation GPU. We evaluated models based on latency, resource consumption, and impact on task success.

\textbf{Metrics.} We evaluated three key performance indicators: 
 \textbf{Average feedback latency}: the mean elapsed time from sensor data arrival to publishing the semantic reasoning result on the ROS topic (milliseconds).
  \textbf{Peak VRAM usage}: the maximum GPU memory consumed by the combined perception and reasoning pipeline (gigabytes).
  \textbf{Task success rate}: the percentage of successful landings within 0.5 m of the target, without collisions or close calls (as defined in Experiment 1).
  
\begin{table}[h!]
\centering
\caption{Performance Benchmark of Quantized LLMs}
\label{tab:exp3_llm_benchmark}
\begin{tabular}{@{}l 
    S[table-format=4.1] 
    S[table-format=1.1] 
    S[table-format=3.1]@{}}
\toprule
\textbf{LLM Model (Quantized)} 
  & \textbf{Avg.\ Latency (ms)} 
  & \textbf{Peak VRAM (GB)} 
  & \textbf{Success Rate (\%)} \\
\midrule
Llama-3.2-1B-Instruct                      & 1453.6  & 1.5 & 94.0 \\
Qwen2.5-1.5B-R1-Distilled        & 1510.2  & 2.3 & 70.0 \\
Qwen2.5–3B                       & 1889.5  & 2.9 & 82.0 \\
Llama-3.2–3B-Instruct                    & 2252.7  & 2.9 & 60.0 \\
\bottomrule
\end{tabular}
\end{table}

\textbf{Results.} Tab.~\ref{tab:exp3_llm_benchmark} summarizes the trade‐offs among our quantized LLMs. The smallest model, \texttt{Llama-3.2-1B-Instruct}, delivers the lowest latency (1453.6 ms) and minimal memory footprint (1.5 GB) while achieving a 94.0\% success rate. Mid-sized models (\texttt{Qwen2.5-1.5B-R1-Distilled} and \texttt{Qwen2.5-3B-Instruct}) show moderate increases in latency and VRAM (1010.2–1266.5 ms; 2.3–2.9 GB) with corresponding drops in success (70.0–82.0\%). The largest model, \texttt{Llama-3.2-3B-Instruct}, exhibits the highest latency (2252.7 ms) and peak VRAM usage (2.9 GB), impairing its real-time responsiveness and yielding only a 60.0\% success rate due to delayed MPC inputs. Conversely, \texttt{Llama-3.2-1B-Instruct} delivers the optimal performance with the fastest inference, lowest resource consumption, and highest landing success on constrained hardware.


%% file: sections/5_conclusion_and_discussion.tex

Ensuring safe and reliable autonomous UAV landing in dynamic environments remains challenging. This paper introduced a novel framework integrating a lightweight, RAG-enhanced LLM directly into an MPC-based local planner for UAV control. Our method leverages semantic reasoning from LLMs to improve obstacle identification, classification, and subsequent landing decisions.

Experimental results confirmed that the proposed LLM-Land significantly outperformed baseline MPC and non-augmented LLM-MPC approaches in dynamic simulated environments. The integration of semantic context notably increased landing safety and success rates. Critically, optimized and quantized lightweight LLMs were demonstrated to run effectively in real-time on resource-constrained hardware such as the NVIDIA Jetson Orin NX, bridging advanced AI reasoning with practical robotic applications.

Future work includes autonomously updating the RAG knowledge base to enable continuous learning from operational experiences and exploring larger, more capable models as computational resources improve. Integration of advanced video understanding models or significantly larger foundation models could further enhance the system's contextual understanding and real-time decision-making capabilities. This research validates the potential of grounded LLM reasoning within UAV control loops, paving the way for safer, context-aware aerial robotics suitable for real-world complexities.

%% file: sections/6_appendix.tex
\section{Appendix}

This appendix provides additional details on the overall framework design, with a particular focus on the prompt strategies used to interface with the large language model (LLM) and how its outputs are integrated into the trajectory planning pipeline. We also describe the hardware architecture used for real-world deployment, along with implementation details of the experimental setup, including communication protocols and system integration considerations.

\subsection{Prompt Design and LLM Behavior}

The performance of lightweight language models in our system is sensitive to prompt structure and clarity. To ensure consistent output, we employ a typed system prompt that explicitly instructs the LLM to extract structured metadata only, avoiding any free-text generation. An example prompt is shown below:

\begin{verbatim}
You are a drone safety metadata extractor. Do NOT read any free text.
From the context, pull exactly two fields:
  • Classification: `yes' or `no'
  • Minimum Altitude: a float (strip `meters')

Context:
{context_str}

Examples:
Context: [Classification: no | Minimum Altitude: 0.0 meters | Text: brick building]
Q: a room with tree on the floor?
A: ```json
{"is_dynamic": "no", "z_min": 0.0}

Return only JSON:
```json
{
  "is_dynamic": "{Classification}",
  "z_min": {Minimum_Altitude}
}
```

\end{verbatim}

Smaller LLMs often struggle to produce valid JSON output without strong prompting and in-context examples. We observed that different models interpret the same prompt with varying degrees of compliance. For instance, the Llama model demonstrated relatively stable behavior, while the Qwen models were more prone to formatting errors or misinterpretation of field names. This may be attributed to Qwen's multilingual training corpus, which could introduce ambiguity in prompt interpretation compared to Llama, which are predominantly trained on English-language data.
\subsection{Pipeline Example}

\begin{figure}[!ht]
    \centering
    \includegraphics[width=\linewidth]{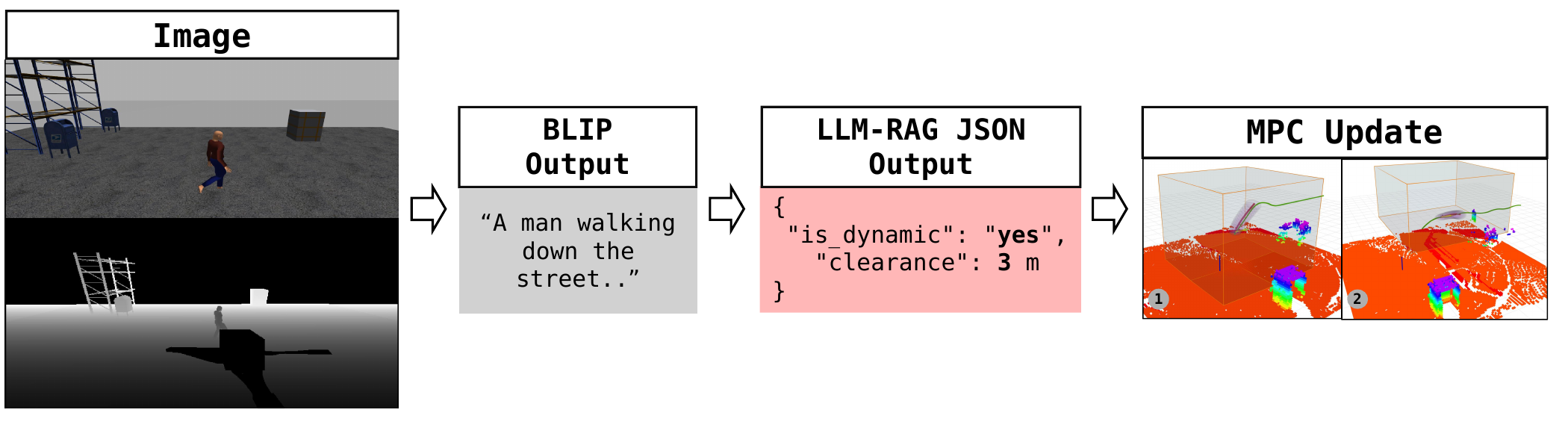}
    \caption{Illustration of the LLM-land pipeline from visual perception to semantic reasoning and trajectory updates.}
    \label{fig:logic}
    \vspace{-0.5cm}
\end{figure}

Figure~\ref{fig:logic} illustrates a complete example of the LLM-land pipeline. In this scenario, the UAV captures an onboard image and sends it to a BLIP-based vision-language encoder, which generates a descriptive caption of the scene. This caption is then passed to the LLM-RAG module, which queries the onboard knowledge base and returns a structured JSON response containing obstacle classification and minimum altitude information. The resulting metadata is forwarded to the MPC controller, which uses it to update the UAV’s trajectory accordingly. Additional examples demonstrating this end-to-end process are provided in the supplementary video material.

\subsection{Unsafe Region Awareness in Planning}

To ensure safe navigation in environments with dynamic obstacles, relying solely on local reactive avoidance within a Model Predictive Control (MPC) framework is often insufficient. 
MPC operates over a limited time horizon and assumes a safe reference trajectory; if this reference intersects unsafe regions, the controller may either violate safety constraints or exhibit unstable behavior. 
To address this limitation, we also incorporate unsafe region awareness at the planning level by shaping the heuristic used during trajectory search. 
By incorporate soft penalties into the heuristic, the planner biases candidate trajectories away from unsafe zones, ensuring that the resulting reference path is both dynamically feasible and safety-aware before being passed to the MPC for tracking.

In kinodynamic trajectory search, the total cost function is composed of an actual cost and a heuristic function: $f(\mathbf{p}) = g(\mathbf{p}) + h_0(\mathbf{p})$
where $g(\mathbf{p})$ is the accumulated cost along a dynamically feasible trajectory, and $h_0(\mathbf{p})$ estimates the minimum effort required to reach the goal from state $\mathbf{p}$.
To encourage spatial safety under dynamic obstacles, we augment the heuristic with a soft penalty for proximity to unsafe regions, using the modified heuristic $\tilde{h}(\mathbf{p}) = h_0(\mathbf{p}) + \lambda \, \phi(\mathbf{p})$.
where $\lambda > 0$ is a weighting parameter, and $\phi(\mathbf{p}) \geq 0$ is the unsafe region penality function $\phi: \mathbb{R}^3 \to \mathbb{R}$ as:
\begin{equation}
    \phi(\mathbf{p}) =
\begin{cases}
0, & \text{if } \mathbf{p} \in \mathcal{B} \\
\delta(\mathbf{p}), & \text{otherwise}.
\end{cases}
\end{equation}
The penalty term $\delta(\mathbf{p})$ increases smoothly as the point $\mathbf{p}$ deviates from these corridors. For instance, under a one-dimensional constraint along axis $d$, the penalty function takes the form:
\begin{equation}
\delta_d(p_d) =
\begin{cases}
(p_d - b_{\min,d})^2 + \epsilon, & \text{if } p_d < b_{\min,d} \\
(p_d - b_{\max,d})^2 + \epsilon, & \text{if } p_d > b_{\max,d} \\
0, & \text{otherwise},
\end{cases}
\end{equation}
where $b_{\min,d}$ and $b_{\max,d}$ define the lower and upper bounds along dimension $d$, and $\epsilon > 0$ ensures numerical stability.
In addition, to prevent overly restrictive behavior near the start and goal states, we define additional clearance regions around both endpoints to expand the safe region.

When extended to two or three dimensions, the unsafe region penalty may require indicator-based formulations to capture its structure precisely, particularly when the safe region is defined as a union of axis-aligned boxes. To avoid the complexity and discontinuities introduced by such formulations, a common and effective approach is to approximate the penalty using a convex lower bound, typically based on the squared distance to the nearest face or boundary of the safe region.
\subsection{Real-World Demonstration}
\begin{figure}[!ht]
    \centering
    \includegraphics[width=0.9\linewidth]{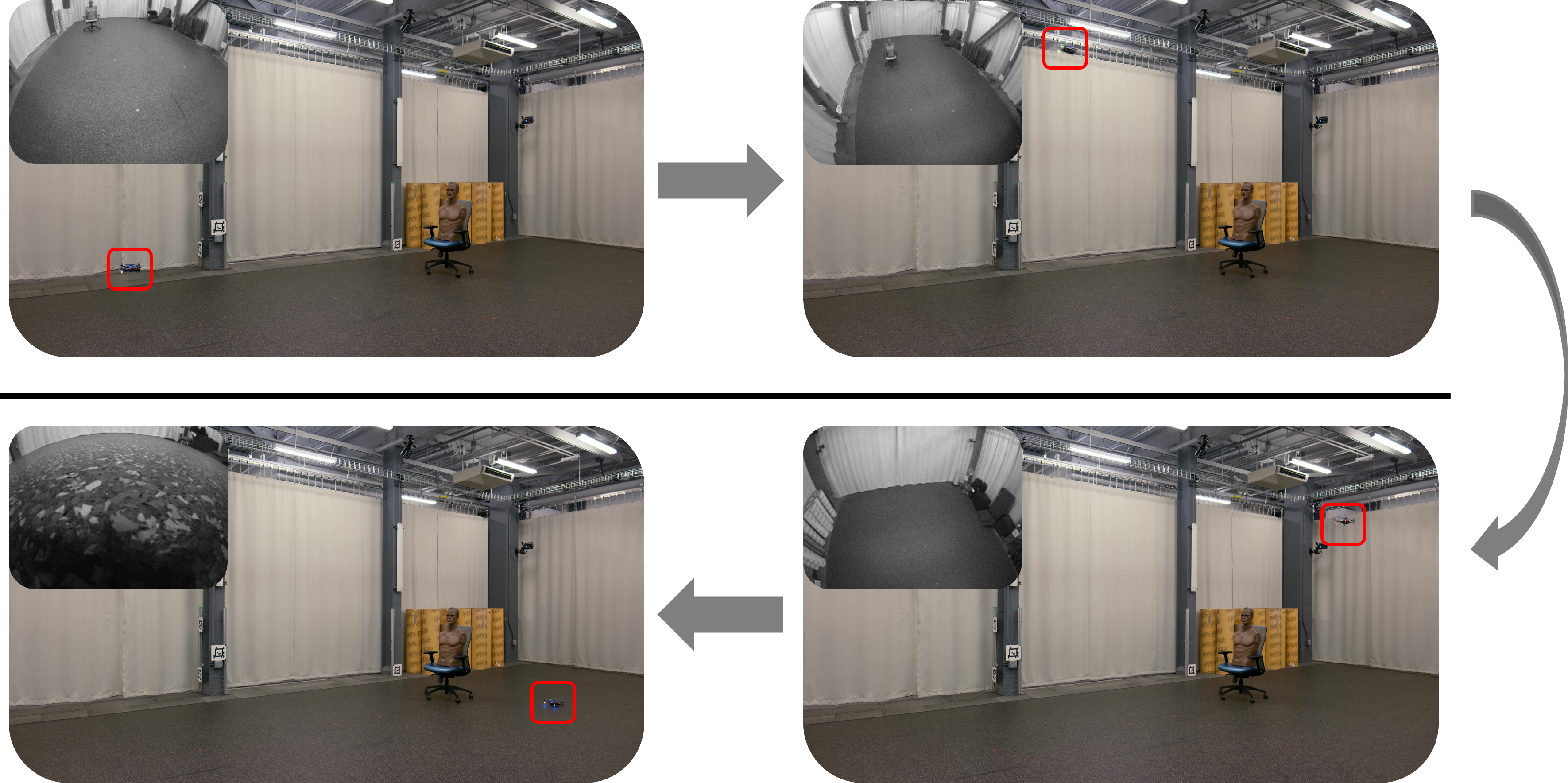}
       \caption{Indoor flight test with the custom-built UAV. The UAV detected a manikin near its path, and the LLM-land module classified it as a dynamic obstacle that needs to be avoided actively. The UAV adjusted its trajectory to fly over the manikin and completed a safe landing.}
    \label{fig:Realworld}
    \vspace{-0.3cm}
\end{figure}

To qualitatively evaluate the feasibility of our system in a physical setting, we conducted indoor flight tests using a custom-built UAV platform.

\begin{figure}[!ht]
    \centering
    \includegraphics[width=0.5\linewidth]{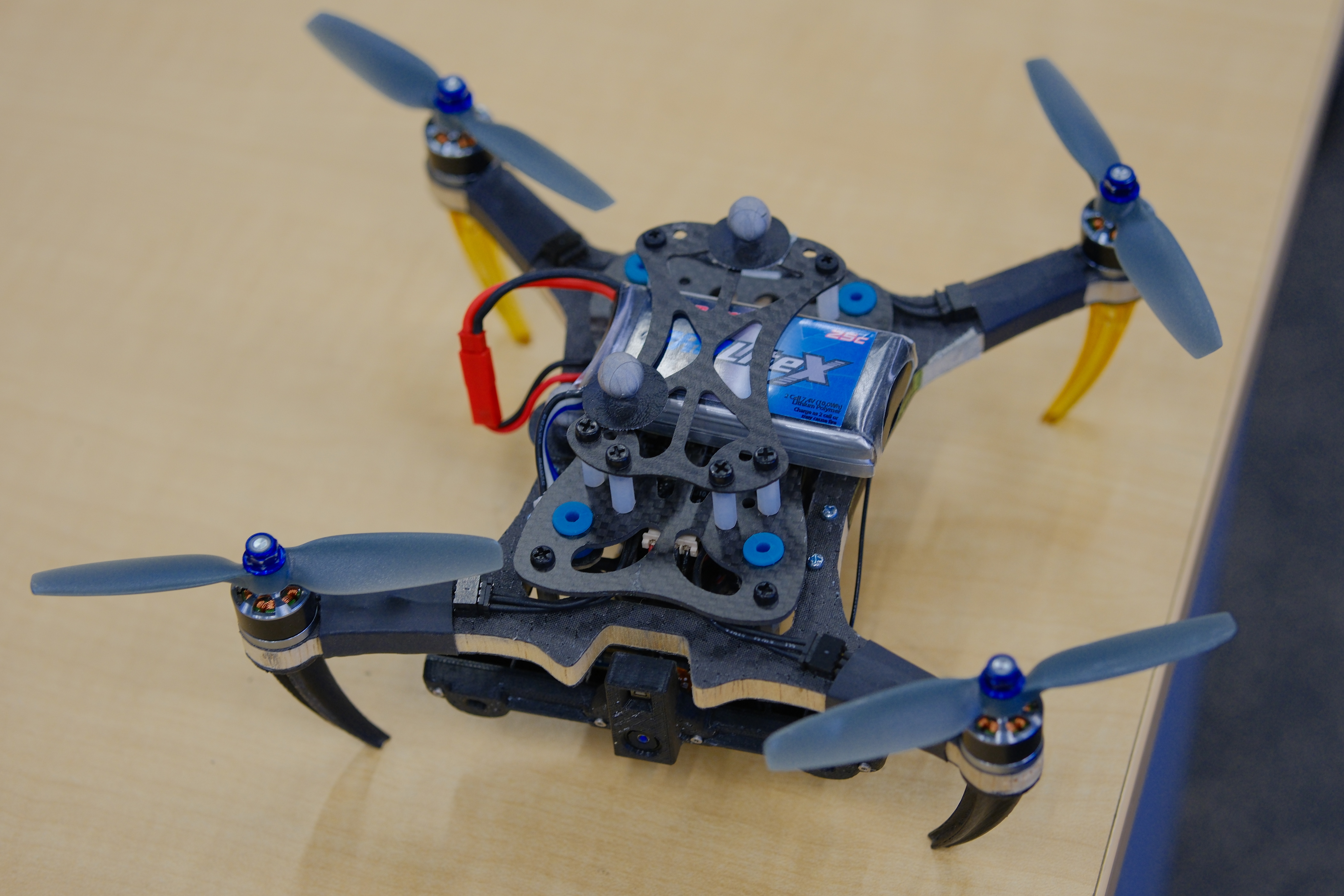}
    \caption{Lightweight UAV platform based on VOXL and PX4.}
    \label{fig:UAV}
    \vspace{-0.3cm}
\end{figure}

\paragraph{Hardware Setup.}  
The UAV platform (Fig.~\ref{fig:UAV}) features a lightweight wooden frame powered by a 2S LiPo battery. It is equipped with a VOXL compute unit, a 45° front-facing monochrome camera, and a Pixhawk flight controller running PX4 firmware. Visual-semantic reasoning is performed offboard on a ground station. The ground station is a laptop configured with an Intel i7-12700K CPU and an NVIDIA RTX 3070 Ti GPU. Communication between the UAV and the ground station is established over WIFI.

\paragraph{Test Environment.}  
Tests were conducted indoors under motion capture supervision using a Vicon tracking system. The environment included a static obstacle, represented by a life-sized manikin seated on a chair, positioned adjacent to the UAV's intended flight path.

\paragraph{Scenario and Outcome.}  
As shown in Fig~\ref{fig:Realworld}, The UAV was instructed to take off and follow a straight-line trajectory toward a designated landing point. Upon detecting the manikin, the LLM-Land module identified it as a potential obstacle. In response, the UAV autonomously altered its path by ascending above the obstacle and then continued to the target location, where it performed a safe landing. These results demonstrate the system’s ability to translate semantic visual input into meaningful control behavior. A video recording of this trial is included in the supplementary materials.